\pdfoutput=1

\documentclass[11pt]{article}

\usepackage[]{acl}

\usepackage{times}
\usepackage{latexsym}

\usepackage[T1]{fontenc}

\usepackage[utf8]{inputenc}

\usepackage{microtype}

\usepackage{inconsolata}

\usepackage{amssymb}
\usepackage{amsthm}
\usepackage{amsmath}
\usepackage{graphicx}
\usepackage{multirow}
\usepackage{booktabs}
\theoremstyle{definition}
\newtheorem{definition}{Definition}
\newtheorem{proposition}{Proposition}

\title{Temporal Knowledge Graph Completion with Time-sensitive Relations in Hypercomplex Space}

\author{Li Cai$^{1,2}$, Xin Mao$^{1}$, Zhihong Wang$^1$, Shangqing Zhao$^1$, Yuhao Zhou$^1$, Changxu Wu$^3$, Man Lan$^{1}$ \thanks{*Corresponding author}\\
$^1$School of Computer Science and Technology, East China Normal University \\
$^2$College of Computer Science and Technology, Guizhou University \\
$^3$Department of Industrial Engineering, Tsinghua University \\
\texttt{\{caili2020, xmao, sqzhao, zhihong, 51265900018\}@stu.ecnu.edu.cn} \\
\texttt{wuchangxu@tsinghua.edu.cn, mlan@cs.ecnu.edu.cn}}

\begin{document}
\maketitle
\begin{abstract}
Temporal knowledge graph completion (TKGC) aims to fill in missing facts within a given temporal knowledge graph at a specific time. Existing methods, operating in real or complex spaces, have demonstrated promising performance in this task. This paper advances beyond conventional approaches by introducing more expressive quaternion representations for TKGC within hypercomplex space. Unlike existing quaternion-based methods, our study focuses on capturing time-sensitive relations rather than time-aware entities. Specifically, we model time-sensitive relations through time-aware rotation and periodic time translation, effectively capturing complex temporal variability. Furthermore, we theoretically demonstrate our method's capability to model symmetric, asymmetric, inverse, compositional, and evolutionary relation patterns. Comprehensive experiments on public datasets validate that our proposed approach achieves state-of-the-art performance in the field of TKGC.
\end{abstract}

\section{Introduction}

Knowledge graphs (KGs) serve as organized repositories of information about the world, utilizing a triplet format \emph{(h, r, t)} to represent factual knowledge. In this format, $h$ denotes the head entity, $r$ represents the relation, and $t$ signifies the tail entity. Due to their ability to capture complex relations, KGs have emerged as essential resources in various applications, including search engines~\cite{DBLP:journals/tnn/ZhaoCXM23SE}, question-answering systems~\cite{DBLP:conf/acl/WangZLL23QA}, and recommendation systems~\cite{DBLP:conf/sigir/ChenGL023RS}.

Despite their significance, knowledge graphs (KGs) inherently lack completeness due to the vastness of real-world information. Knowledge graph completion (KGC) becomes crucial for inferring absent relations or entities within KGs.

While KGs store static facts about the world, real-world data is dynamic, with facts evolving over time. This necessitates an extension of traditional KGs to incorporate temporal dimensions, giving rise to temporal knowledge graphs (TKGs). Within TKGs, facts are represented as quadruples $(h, r, t, \tau)$,  where $\tau$ is the timestamp.

\begin{figure}
  \centering  
  \includegraphics[width=0.99\linewidth]{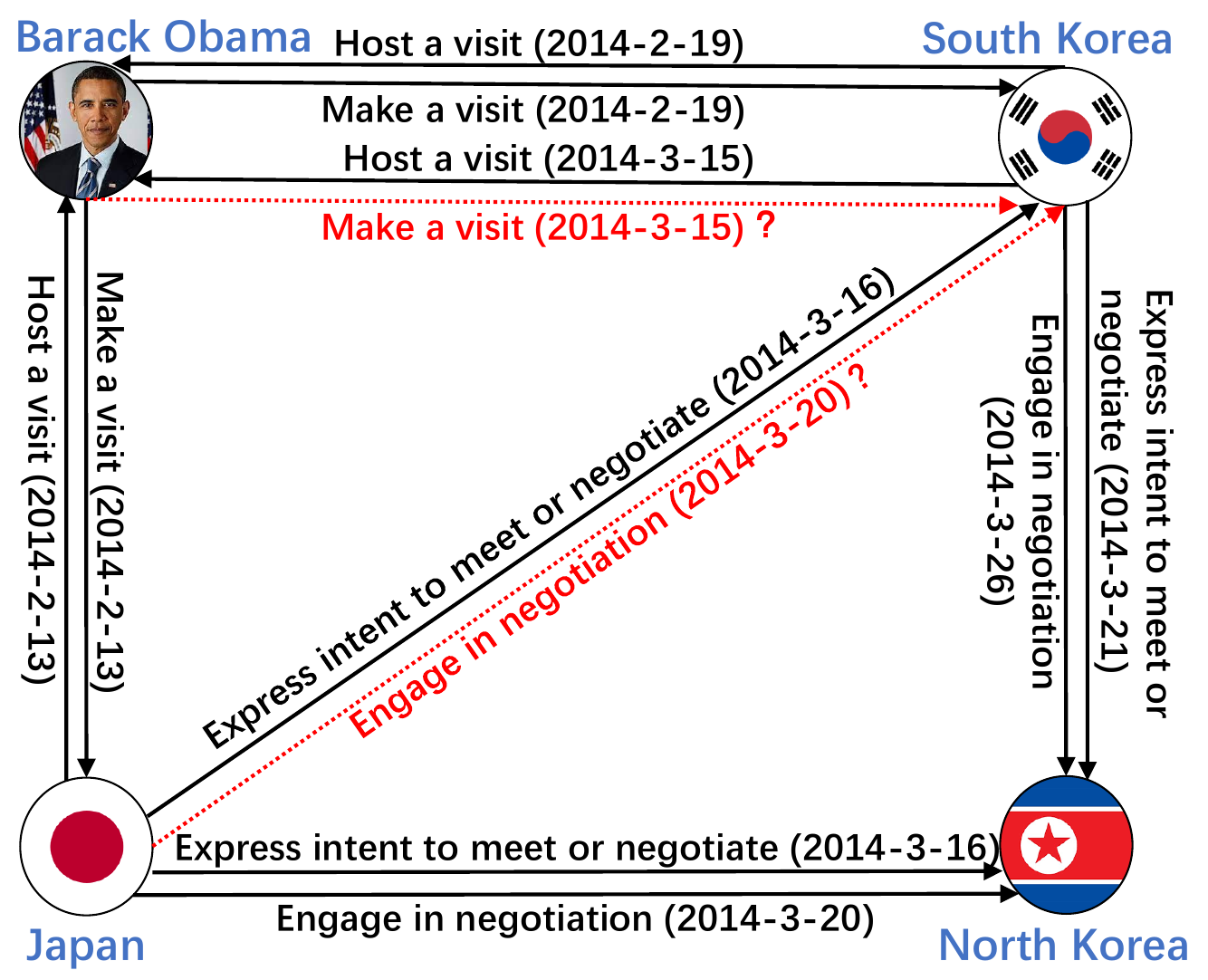}
  \caption{Subgraph example from ICEWS14 dataset for TKGC. Nodes on the graph represent entities, including {Obama, Japan, South Korea, North Korea}, and edges signify relations (with timestamps in parentheses), encompassing {Host a visit, Make a visit, Express intent to meet or negotiate, Engage in negotiation}. Solid black arrows denote existing facts, while red dashed arrows indicate missing facts that need to be completed.}
  \label{fig:example}
\end{figure}

Temporal knowledge graph completion (TKGC) naturally extends the traditional KGC task to address temporal aspects of evolving facts. In Figure~\ref{fig:example}, a case of TKGC is illustrated using a subgraph of ICEWS14. This subgraph contains facts like \emph{(South Korea, Host a visit, Barack Obama, 2014-2-19), (Barack Obama, Make a visit, South Korea, 2014-2-19), (South Korea, Host a visit, Barack Obama, 2014-3-15)}. TKGC aims to explore the temporal features and periodic patterns associated with these facts and subsequently complete missing facts, such as \emph{(Barack Obama, Make a visit, South Korea, 2014-3-15)}.

Existing TKGC methodologies build upon established KGC paradigms and consider the temporal dynamics of entities or relations over time, thus achieving superior performance compared to conventional KGC methods. TTransE~\cite{10.1145/3184558.3191639-TTransE}, TA-TransE~\cite{Garcia-DuranDN18tatranse}, and HyTE~\cite{dasgupta18hyte} are translation-based models that learn the embeddings of entities, relations, and times in real space.
Tero~\cite{xu20tero}, TComplEx~\cite{LacroixOU20TComplEx}, TeLM~\cite{DBLP:conf/naacl/XuCNL21TeLM}, and TeAST~\cite{DBLP:conf/acl/LiSG23TeAST} extend learned representations into complex space, thereby further enhancing the performance of TKGC. RotateQVS~\cite{chen-etal-2022-rotateqvs} and TLT-KGE~\cite{10.1145/3511808.3557233-TLTKGE} further extend representations into hypercomplex space with quaternions, indicating that hypercomplex space possesses better representational capabilities than complex space. 

Although methods based on hypercomplex space for TKGC are successful, they have their shortcomings. RotateQVS first rotates the entities to the temporal dimension to get the time-aware entities, then posit relation as a translation from the time-aware head entity to the time-aware tail entity. It falls short in exploring the temporal variations of relations over time.  
TLT-KGE integrates temporal representation into the representation of entities and relations, combining their semantic representation in the real part with the temporal representation in the imaginary part. It also introduces a Shared Time Window (STW) module to enhance the associations between events occurring within a specific time period. TLT-KGE offers an improved capability to model the evolution of knowledge graphs over time. However, since the size of time windows is a hyper-parameter and varies across different datasets, extensive experimentation is required to determine the optimal value, leading to inefficiency.  

Therefore, this paper explores the hypercomplex space for learning TKG embeddings. Specifically, we use quaternion embeddings to represent entities, relations and times. Each quaternion embedding is a vector in the hypercomplex space $\mathbb{H}$ with one real component $a$ and three imaginary components $b\mathbf{i}$, $c\mathbf{j}$, and $d\mathbf{k}$. We employ time-aware rotation and periodic time translation to model time-sensitive relations, enhancing our capability to capture the complex temporal variability of facts. The time-aware rotation utilizes the Hamilton product of quaternions to capture local temporal interactions, while periodic time translation captures interactions within a time period. Additionally, we provide theoretical evidence demonstrating that our approach covers symmetric, asymmetric, inverse, compositional, and evolutionary relation patterns.
Experimental results on public datasets, including ICEWS14, ICEWS05-15, and GDELT, demonstrate that our model achieves state-of-the-art (SOTA) performance. Notably, our model exhibits remarkable performance improvements on GDELT, which is known for its complex temporal variability. Compared to TLT-KGE, the improvement scores for MRR, Hits@1, Hits@3, and Hits@10 on GDELT are 8.38\%, 13.96\%, 7.47\%, and 2.03\%, respectively.

The primary contributions of this paper can be summarized as follows:

(1) We propose a novel approach to represent TKGs in hypercomplex space. By utilizing quaternion embeddings and incorporating time-aware rotation and periodic time translation, our method effectively captures the complex temporal variability of facts within TKGs.

(2) Theoretical analyses and proofs are provided to demonstrate the capability of our proposed method in modeling various relation patterns, including symmetric, asymmetric, inverse, compositional, and evolutionary relation patterns.

(3) Comprehensive experiments on various public datasets demonstrate that our approach achieves SOTA performance. 

\section{Related Works}

\subsection{Knowledge Graph Completion}

Traditional KGC, often referred to as link prediction, typically involves learning embeddings for head entities, relations , and tail entities, denoted as $\mathbf{e}_h, \mathbf{e}_r, \mathbf{e}_t$. This methodology aims to unveil intricate patterns of relations between entities, thereby facilitating the completion of missing facts.

TransE~\cite{DBLP:conf/nips/BordesUGWY13TransE} is a foundational model in KGC, treating relation embedding as a translation from head entity embedding to tail entity embedding, represented as $\mathbf{e}_h + \mathbf{e}_r \approx \mathbf{e}_t$. Subsequently, a series of translation-based models emerged as derivatives, including TransD, TransH, TransR, and others. These models represent embeddings in the real space $\mathbb{R}$, yet may lack the ability to represent varied relation patterns.

ComplEx~\cite{DBLP:conf/icml/TrouillonWRGB16ComplEx} and RotatE~\cite{DBLP:conf/iclr/SunDNT19-rotate} extend embeddings from the real space $\mathbb{R}$ to the complex space $\mathbb{C}$. ComplEx utilizes a bilinear score function $Re(\left\langle \mathbf{e}_h,\mathbf{e}_r,\mathbf{\overline{e}}_t \right\rangle)$ to assess compatibility between entities and relations in the KG.
On the other hand, RotatE views the relation embedding as a rotation from head entity embedding to tail entity embedding, denoted as $\mathbf{e}_h \circ \mathbf{e}_r \approx \mathbf{e}_t$. These models enhance the capability to model relationship patterns by leveraging complex numbers.

QuatE~\cite{DBLP:conf/nips/0007TYL19QuatE} and QuatRE~\cite{DBLP:conf/www/NguyenVNP22QuatRE} go beyond complex representations, exploring hypercomplex space $\mathbb{H}$ for learning KG embeddings. Quaternion, with its enhanced representation capability, enables the model to capture intricate interactions and dependencies within a KG. QuatE utilizes the score function $\mathbf{e}_h \otimes \mathbf{e}_r^\triangleleft \cdot \mathbf{e}_t$, and QautRE utilizes the score function $(\mathbf{e}_h \otimes (\mathbf{e}_{r,1}^\triangleleft \otimes \mathbf{e}_{r}^\triangleleft)) \cdot (\mathbf{e}_t \otimes \mathbf{e}_{r,2}^\triangleleft)$. Both of these models can effectively capture more flexible relation patterns.

Despite the significant advancements brought by these methods in KGC, they share a common limitation of neglecting temporal information in KGs.

\subsection{Temporal Knowledge Graph Completion}

TKGC is an extension of traditional KGC methods that incorporate temporal information to capture the dynamic evolution of facts in a TKG over time. Unlike traditional KGC, which focuses on static representations, TKGC acknowledges the temporal dimension, recognizing that relations between entities can change over different time intervals.

TTransE~\citep{10.1145/3184558.3191639-TTransE} incorporates temporal information by concatenating it with relations and employs the score function $\Vert\mathbf{e}_h+\mathbf{e}_r+\mathbf{e}_\tau-\mathbf{e}_t\Vert$.
TA-TransE~\citep{Garcia-DuranDN18tatranse}  represents relations as sequences with temporal information and learns the sequences embedding as the translation from the head entity to the tail entity.
HyTE~\citep{dasgupta18hyte} projects entities and relations onto the time hyperplane and regards the projected relations as the translation of the projected head entity to the projected tail entity.
However, all these methods, based on translation in real space, face challenges similar to translation-based methods in KGC.

TComplEx~\citep{LacroixOU20TComplEx} builds upon ComplEx~\citep{DBLP:conf/icml/TrouillonWRGB16ComplEx} by incorporating temporal embeddings and introducing new regularization schemes.
Tero~\citep{xu20tero} characterizes the temporal evolution of entity embeddings by representing rotations from their initial time to the current time in complex space. 
TeLM~\cite{DBLP:conf/naacl/XuCNL21TeLM} conducts 4-order tensor factorization on a TKG, employing linear temporal regularization and multivector embeddings.
ChronoR~\cite{sadeghian2021chronor} considers relations and timestamps as rotations of the head entity in a k-dimensional space. 
BoxTE~\cite{DBLP:conf/aaai/MessnerAC22BoxTE} extends the BoxE~\cite{DBLP:conf/nips/AbboudCLS20boxe} for TKGC by incorporating box embeddings.
TeAST~\cite{DBLP:conf/acl/LiSG23TeAST} maps relations to an Archimedean spiral timeline and transforms quadruple completion into a 3rd-order tensor Decomposition problem. 

RotateQVS~\citep{chen-etal-2022-rotateqvs} utilizes a hypercomplex space to represent entities, relations, and timestamps. Temporal entity representations are acquired through temporal rotations in quaternion vector space.
TLT-KGE~\citep{10.1145/3511808.3557233-TLTKGE} incorporates time representation into the entity and relation embeddings. These approaches highlight the increased expressiveness of hypercomplex space compared to complex space.

Quaternions offer greater expressive power than complex numbers thanks to Hamilton multiplication, facilitating more comprehensive interactions. Hence, exploring the potential of quaternions in TKGC is essential.

\section{Problem Formulation and Quaternion}
\subsection{Problem Formulation of TKGC}
A TKG is represented as a directed graph $\mathcal{G=(E,R,T,F)}$, where $\mathcal{E}$, $\mathcal{R}$ and $\mathcal{T}$ represents the sets of entities, relations and timestamps respectively.
The set $\mathcal{F\subset E\times R\times E\times T}$ comprises facts in the form $(h,r,t,\tau)$,
where $h \in \mathcal{E}$ presents the head entity, $r \in \mathcal{R}$ presents the relation, $t \in \mathcal{E}$ presents the tail entity, and $\tau \in \mathcal{T}$ presents the timestamp. For a given TKG $\mathcal{G}$ and a query $(h,r,?,\tau)$ or $(?,r,t,\tau)$, TKGC aims to predict and complete the missing entity in the query.

\subsection{Quaternion Algebra}
A quaternion is a hypercomplex that extends the concept of complex numbers. It was first introduced by the Irish mathematician William Rowan Hamilton in 1843~\cite{hamilton1843quaternions}. Unlike complex numbers, which have two components (one real part and one imaginary part), quaternions have four components: one real part and three imaginary parts. 

The general form of a quaternion is expressed as:
$ q = a+b \mathbf {i} +c \mathbf {j} +d \mathbf {k}$, where $q \in \mathbb{H}$, $ a, b, c, d \in \mathbb {R} $, and $ \mathbf {i}, \mathbf {j}, \mathbf {k} $ are three imaginary parts that satisfy $ \mathbf {i}^2 = \mathbf {j}^2 = \mathbf {k}^2 = \mathbf {ijk} = {-1} $. From which we can derive $\mathbf {ij} = \mathbf {-ji} = \mathbf {k} ,\  \mathbf {jk} = \mathbf {-kj} = \mathbf {i} ,\  \mathbf {ki} = \mathbf {-ik} = \mathbf {j}$.

Below are fundamental calculations and properties related to quaternions:

\noindent \textbf{Hamilton Product (Quaternion Multiplication):} The multiplication of quaternions is their fundamental algebraic operation. Given two quaternions $q_1 = a_1 + b_1\mathbf{i} + c_1\mathbf{j} + d_1\mathbf{k}$ and $q_2 = a_2 + b_2\mathbf{i} + c_2\mathbf{j} + d_2\mathbf{k}$, their product $q_1 \otimes q_2$ can be computed as follows:
\begin{equation}
\begin{aligned}
q_1 \otimes q_2 & =\left(a_1 a_2-b_1 b_2-c_1 c_2-d_1 d_2\right) \\
& +\left(a_1 b_2+b_1 a_2+c_1 d_2-d_1 c_2\right) \mathbf{i} \\
& +\left(a_1 c_2-b_1 d_2+c_1 a_2+d_1 b_2\right) \mathbf{j} \\
& +\left(a_1 d_2+b_1 c_2-c_1 b_2+d_1 a_2\right) \mathbf{k}
\end{aligned}
\label{eq1}
\end{equation}

\noindent \textbf{Inner Product:} the inner product (also known as the dot product) of two quaternions is computed as follows:
\begin{equation}
q_1 \cdot q_2 = a_1a_2 + b_1b_2 + c_1c_2 + d_1d_2
\label{eq2}
\end{equation}

\noindent \textbf{Conjugate:} The conjugate $\overline{q}$ of a quaternion $q = a + b\mathbf{i} + c\mathbf{j} + d\mathbf{k}$ is obtained by negating its imaginary components:
\begin{equation}
\overline{q} = a - b\mathbf{i} - c\mathbf{j} - d\mathbf{k}
\label{eq3}
\end{equation}

\section{Methods}
The TKGC model consists of four fundamental components: temporal representation, score function, regularizer, and loss function. 

\subsection{Temporal Reprenstation}
We use quaternions to represent the embeddings of head entities, relations, tail entities, and timestamps, specifically. 
\begin{equation}   
\begin{aligned}
&\mathbf{q}_h = \mathbf{e}^{a}_{h} + \mathbf{e}^{b}_{h}\mathbf{i} + \mathbf{e}^{c}_{h}\mathbf{j} + \mathbf{e}^{d}_{h}\mathbf{k}\\
&\mathbf{q}_r = \mathbf{e}^{a}_{r} + \mathbf{e}^{b}_{r}\mathbf{i} + \mathbf{e}^{c}_{r}\mathbf{j} + \mathbf{e}^{d}_{r}\mathbf{k}\\
&\mathbf{q}_t = \mathbf{e}^{a}_{t} + \mathbf{e}^{b}_{t}\mathbf{i} + \mathbf{e}^{c}_{t}\mathbf{j} + \mathbf{e}^{d}_{t}\mathbf{k}\\
&\mathbf{q}_\tau = \mathbf{e}^{a}_{\tau} + \mathbf{e}^{b}_{\tau}\mathbf{i} + \mathbf{e}^{c}_{\tau}\mathbf{j} + \mathbf{e}^{d}_{\tau}\mathbf{k}\\
\label{eq4}  
\end{aligned}
\end{equation}   
where $\mathbf{q}_h,\mathbf{q}_r,\mathbf{q}_t,\mathbf{q}_\tau\in \mathbb{H}^d$, $\mathbf{e}^*_a,\mathbf{e}^*_b,\mathbf{e}^*_c,\mathbf{e}^*_d \in \mathbb{R}^d$. $d$ is the dimension of embeddings, $*$ represents $h,r,t,\tau$.

In TKGs, relations exhibit temporal and periodic characteristics. This signifies that relations in the TKG not only change over time but may also adhere to certain periodic patterns. Temporality indicates that relation changes are correlated with time, while periodicity suggests that such changes may exhibit repetitive patterns over time. Therefore, we consider modeling relations that evolve over time rather than entities.

Inspired by RotateQVS~\cite{chen-etal-2022-rotateqvs}, we represent temporal relations as rotations in quaternion space. For each time embedding $\mathbf{q}_\tau$ (constrained as a unit quaternion), the mapping is a rotation in quaternion space from the base relation embedding $\mathbf{q}_r$ to the time-aware relation embedding $\mathbf{q}_{r(\tau)}$, defined as follows:
\begin{equation}    
   \mathbf{q}_{r(\tau)} = \mathbf{q}_\tau \otimes \mathbf{q}_r \otimes \mathbf{\overline{q}}_\tau \\
   \label{eq5}       
\end{equation}
where $\otimes$ means Hamilton product. The multiplication of quaternions facilitates more interactions between time and relations.

In recent TKGR models~\cite{DBLP:conf/acl/LiSG23TeAST,DBLP:conf/sigir/0006MLLTWZL23rpc}, the periodicity of time is commonly modeled using the sine function. Inspired by this, we adopt the following approach to model the periodicity of relationships. First, we use a quaternion vector $\mathbf{q}_{\tau'} = \mathbf{e}^{a}_{\tau'} + \mathbf{e}^{b}_{\tau'}\mathbf{i} + \mathbf{e}^{c}_{\tau'}\mathbf{j} + \mathbf{e}^{d}_{\tau'}\mathbf{k}\\$ to represent periodic time. 
Then, we calculate the sin value of the periodic time. We add this sine value to the time-aware relation to obtain the time-sensitive relation $\mathbf{q'}_{r(\tau)}$. This representation captures both the temporal and periodic features of the relation. The process can be described using the following formula:
\begin{equation}
  \begin{aligned}
    \mathbf{q'}_{r(\tau)} & = \mathbf{q}_{r(\tau)} + sin(\mathbf{q}_{\tau'}) \\
    & = \mathbf{q}_\tau \otimes \mathbf{q}_r \otimes \mathbf{\overline{q}}_\tau + sin(\mathbf{q}_{\tau'}) \\
    \label{eq6}
  \end{aligned} 
\end{equation}

\subsection{Score Function}

Previous successful studies~\cite{DBLP:conf/icdt/AggarwalHK01-distance1,DBLP:journals/sadm/ZimekSK12-distance2} have demonstrated that cosine distance often offers a more effective measure of similarity between vectors compared to translation distance.

Different from RotateQVS~\cite{chen-etal-2022-rotateqvs}, which uses the Euclidean distance for measuring vector similarity, we adhere to the approach of QuatE~\cite{DBLP:conf/nips/0007TYL19QuatE} and utilize the cosine distance between two vectors to indicate their similarity.

As a result, we employ the inner product of two vectors to represent the cosine similarity between $\mathbf{q}_{h(r,\tau)}$ and $\mathbf{q}_t$, and the score function is as follows:

\begin{equation}
\begin{aligned}
    \phi(h,r,t,\tau) & = \mathbf{q}_{h(r,\tau)} \cdot \mathbf{q}_t \\
    & = \mathbf{q}_{h} \otimes \mathbf{q'}_{r(\tau)} \cdot \mathbf{q}_t
    \label{eq7}
\end{aligned}
\end{equation}

\subsection{Regularizer}

Regularizers have been employed in prior studies to improve a model's ability to generalize to unseen facts and reduce overfitting to the training data. These regularizers typically fall into two categories: embedding regularizers and temporal regularizers.

Following the approach of TLT-KGE~\cite{10.1145/3511808.3557233-TLTKGE} and TeAST~\cite{DBLP:conf/acl/LiSG23TeAST}, the embedding regularizer $\Omega$ of TQuatE is expressed as:
\begin{equation}
   \Omega = \sum_{i=1}^{k}(\|\mathbf{q}_h\|_p^p + \| \mathbf{q'}_{r(\tau)}\|_p^p + \| \mathbf{q}_t\|_p^p) 
   \label{eq8}
\end{equation}
where $\| \cdot \|_p^p$ denotes the $p$-norm operation for vectors.

The temporal regularizer constrains the temporal embeddings, enhancing the modeling of TKGs. TComplEx~\cite{LacroixOU20TComplEx} introduced smooth temporal regularization in TKGC, aiming for neighboring timestamps to possess similar representations.
TeLM~\cite{DBLP:conf/naacl/XuCNL21TeLM} introduces a linear temporal regularizer that incorporates a bias component between adjacent temporal embeddings.
TLT-KGE~\cite{10.1145/3511808.3557233-TLTKGE} incorporates the shared time window embedding into the smoothing temporal regularizer, ensuring that adjacent timestamps and those within the shared time window have similar embedding.
TeAST~\cite{DBLP:conf/acl/LiSG23TeAST} proposes a temporal spiral regularizer by introducing the phase timestamp embedding into the smoothing temporal regularizer.
In this paper, we enhance the smoothing temporal regularizer by incorporating the periodic time embedding. The resulting periodic temporal regularizer $\Lambda$ is formulated as follows:
\begin{equation}
\begin{aligned}
   \Lambda = \frac{1}{(|\mathcal{T}|-1)} & \sum_{i=1}^{\mathcal{T}-1}(\|\mathbf{q}_{\tau(i+1)}-\mathbf{q}_{\tau(i)} \\
   & + \mathbf{q}_{\tau'(i+1)}-\mathbf{q}_{\tau'(i)} \|_p^p)
   \label{eq9}
\end{aligned}
\end{equation}

\subsection{Loss Function}

We adopt the multi-class log-loss in~\cite{DBLP:conf/acl/LiSG23TeAST}, which has demonstrated effectiveness for both KGE models~\cite{DBLP:conf/icml/TrouillonWRGB16ComplEx} and TKGE models~\cite{LacroixOU20TComplEx,DBLP:conf/naacl/XuCNL21TeLM,10.1145/3511808.3557233-TLTKGE}. Specifically, our loss function $\mathcal{L}_c$ is formulated as follows:
\begin{equation}
\begin{aligned}
   \mathcal{L}_c = & -\log(\frac{\exp(\phi(h, r, t, \tau))}{\sum_{h'\in E} \exp(\phi(h', r, t, \tau))})\\
   & -\log(\frac{\exp(\phi(t, r^{-1}, h, \tau))}{\sum_{t'\in E} \exp(\phi(t', r^{-1}, h, \tau))})
   \label{eq10}
\end{aligned}
\end{equation}
where $s'$ is the subject of the negative quadruple $(h', r, t, \tau)$, $o'$ is the object of the negative quadruple $(h, r, t', \tau)$, and $r^{-1}$ is the inverse relation.

The final loss $\mathcal{L}$ is the combination of $\mathcal{L}_c$, embedding regularizer $\Omega$, and periodic temporal regularizer $\Lambda$:
\begin{equation}
\begin{aligned}
   \mathcal{L} = \mathcal{L}_c + \lambda_e\Omega + \lambda_{\tau}\Lambda
   \label{eq11}
\end{aligned}
\end{equation}
where $\lambda_e$ and
$\lambda_{\tau}$ are the weight of regularier $\Omega$ and $\Lambda$, respectively.

\begin{table*}[ht]
\centering
\resizebox{\textwidth}{!}{
  \begin{tabular}
  {p{2.2cm}cccccccp{3.3cm}}
    \toprule
    Datasets & $|\mathcal{E}|$ & $|\mathcal{R}|$ & $|\mathcal{T}|$ & $|Train|$ & $|Valid|$ & $|Test|$ & $Granularity$ & $Time span$\\
    \midrule
    \textbf{ICEWS14} & 7,128 & 230 & 365 & 72,826 & 8,941 & 8,963 & Daily & 2014.1.1-2014.12.31\\
    \textbf{ICEWS05-15}  & 10,488 & 251 & 4,017 & 368,962 & 46,275 & 46,092 & Daily & 2005.1.1-2015.12.31\\
    \textbf{GDELT} & 500 & 20 & 366 & 2,735,685 & 341,961 & 341,961 & Daily & 2015.4.1-2016.3.31\\
  \bottomrule
\end{tabular}}
\caption{Statistics of datasets for temporal knowledge graph completion. $|\mathcal{E}|$, $|\mathcal{R}|$, $|\mathcal{T}|$, $|Train|$, $|Valid|$, and $|Test|$ denote the number of entities, relations, timestamps, facts in training sets, validation sets, and test sets, respectively.}
\label{tab:datasets}
\end{table*}

\subsection{Relation pattern}

In previous TKGC models~\cite{DBLP:conf/iclr/SunDNT19-rotate,DBLP:conf/acl/LiSG23TeAST}, five types of relation patterns have been proposed, and their definitions are as follows:

\begin{definition}
\emph{A relation $r$ is \textbf{symmetric}, if $\forall h, t, \tau, r(h, t, \tau) \land r(t, h, \tau)$ hold True.}
\label{def1}
\end{definition}

\begin{definition}
\emph{A relation $r$ is \textbf{asymmetric}, if $\forall h, t, \tau, r(h, t, \tau) \land \lnot r(t, h, \tau)$ hold True.}
\label{def2}
\end{definition}

\begin{definition}
\emph{A relation $r_1$ is the \textbf{inverse} of relation $r_2$, if $\forall h, t, \tau, r_1(h, t, \tau) \land r_2(t, h, \tau)$ hold True.}
\label{def3}
\end{definition}

\begin{definition}
\emph{A relation $r_1$ is the \textbf{composition} of relation $r_2$ and relation $r_3$, if $\forall x, y, z, \tau, r_2(x, y, \tau) \land r_3(y, z, \tau) => r_1(x, z, \tau)$ hold True.}
\label{def4}
\end{definition}

\begin{definition}
\emph{A relation $r_1$ and $r_2$ are \textbf{evolving} over time from timestamp $\tau_1$ to timestamp $\tau_2$, if $\forall h, t, \tau, r_1(h, t, \tau_1) \land r_2(t, h, \tau_2)$ hold True.}
\label{def5}
\end{definition}

TQuatE is capable of modeling the aforementioned relation patterns. All propositions are enumerated as the following, and the detailed proofs are presented in the Appendix.

\begin{proposition}
\emph{TQuatE can model the symmetric relation pattern. (See proof in Appendix~\ref{sec:appendix A})}
\end{proposition}

\begin{proposition}
\emph{TQuatE can model the asymmetric relation pattern. (See proof in Appendix~\ref{sec:appendix B})}
\end{proposition}

\begin{proposition}
\emph{TQuatE can model the inverse relation pattern. (See proof in Appendix~\ref{sec:appendix C})}
\end{proposition}

\begin{proposition}
\emph{TQuatE can model the  compositional relation pattern. (See proof in Appendix~\ref{sec:appendix D})}
\end{proposition}

\begin{proposition}
\emph{TQuatE can model the evolutionary relation pattern. (See proof in Appendix~\ref{sec:appendix E})}
\end{proposition}

\section{Experiments}

We conduct the experiments on a workstation with a GeForce
RTX 3090 GPU with 28GB memory. The codes and datasets will be
available on GitHub.
\subsection{Datasets and Baselines}
\subsubsection{Datasets}
We conducted experiments on several datasets to evaluate the performance of our proposed method:

\textbf{ICEWS14 and ICEWS05-15} These datasets are subsets extracted from the Integrated Crisis Early Warning System (ICEWS)~\cite{DVN/28117_2015-icews}\footnote{Further details are available at: http://www.icews.com}. ICEWS processes vast amounts of data from digitized news, social media, and other sources to predict, track, and respond to global events, primarily for early warning purposes. ICEWS14 contains events from the year 2014, while ICEWS05-15 covers events occurring between 2005 and 2015.

\textbf{GDELT} This dataset is a subset of the larger Global Database of Events, Language, and Tone (GDELT)~\cite{leetaru2013gdelt}. GDELT is a comprehensive global database that includes broadcast, print, and web news from every country. The GDELT subset used in our experiments contains facts with daily timestamps between April 1, 2015, and March 31, 2016.

The datasets ICEWS14, ICEWS05-15, and GDELT, utilized in this study, are obtained from TeAST~\cite{DBLP:conf/acl/LiSG23TeAST}\footnote{https://github.com/dellixx/TeAST}. In these datasets, the temporal information is presented in the form of time points, such as [2014-03-14]. Due to the large number of quadruples (about $2M$) in the GDELT dataset but only a small number of entities ($500$) and relations ($20$), it is more challenging to model its temporal evolution compared to the ICEWS datasets. The statistics of these datasets are listed in Table~\ref{tab:datasets}.

\subsubsection{Baselines}
In the experiments, we compared our proposed model with two categories of advanced TKGC methods:

\textbf{Static methods}: 

TransE~\cite{DBLP:conf/nips/BordesUGWY13TransE}, ComplEx~\cite{DBLP:conf/icml/TrouillonWRGB16ComplEx}, RotatE~\cite{DBLP:conf/iclr/SunDNT19-rotate}, QuatE~\cite{DBLP:conf/nips/0007TYL19QuatE}.

\textbf{Temporal methods}: 

(1) Real space: TTransE~\cite{10.1145/3184558.3191639-TTransE}, TA-TransE~\cite{Garcia-DuranDN18tatranse}, HyTE~\cite{dasgupta18hyte}, ChronoR~\cite{sadeghian2021chronor}, BoxTE~\cite{DBLP:conf/aaai/MessnerAC22BoxTE};

(2) Complex space:
TComplEx~\cite{LacroixOU20TComplEx}, Tero~\cite{xu20tero}, TeLM~\cite{DBLP:conf/naacl/XuCNL21TeLM}, TLT-KGE(C)~\cite{10.1145/3511808.3557233-TLTKGE}, TeAST~\cite{DBLP:conf/acl/LiSG23TeAST};

(3) Hypercomplex space:
RotateQVS~\cite{chen-etal-2022-rotateqvs}, TLT-KGE(Q)~\cite{10.1145/3511808.3557233-TLTKGE}. 

\begin{table*}
\centering
\resizebox{\textwidth}{!}{
\begin{tabular}{lcccccccccccc}
\hline
\multirow{2}{*}{Models} & \multicolumn{4}{c}{\bf ICEWS14} & \multicolumn{4}{c}{\bf ICEWS05-15} & \multicolumn{4}{c}{\bf GDELT} \\
\cmidrule(rl){2-5}
\cmidrule(rl){6-9}
\cmidrule(rl){10-13}
& {MRR} & {Hits@1} & {Hits@3} & {Hits@10} & {MRR} & {Hits@1} & {Hits@3} & {Hits@10} & {MRR} & {Hits@1} & {Hits@3} & {Hits@10}\\
\hline
TransE$^{*}$ & .280 & .094 & - & .637 & .294 & .090 & - & .663 & - & - & - & -\\
ComplEx$^{*}$ & .467 & .347 & .527 & .716 & .481 & .362 & .535 & .729 & - & - & - & -\\
RotatE$^{*}$ & .418 & .291 & .478 & .690 & .304 & .164 & .355 & .595 & - & - & - & -\\
QuatE$^{*}$ & .471 & .353 & .530 & .712 & .482 & .370 & .529 & .727 & - & - & - & -\\
\hline
TTransE$^{\diamond}$ & .255 & .074 & - & .601 & .271 & .084 & - & .616 & .115 & .000 & .160 & .318\\
TA-TransE$^{*}$ & .275 & .095 & - & .625 & .299 & .096 & - & .668 & - & - & - & -\\
HyTE$^{\diamond}$ & .297 & .108 & .416 & .655 & .316 & .116 & .445 & .681 & .118 & .000 & .165 & .326\\
ChronoR$^{\dagger}$ & .625 & .547 & .669 & .773 & .675 & .596 & .723 & .820 & - & - & - & -\\
BoxTE$^{\dagger}$ & .613 & .528 & .664 & .763 & .667 & .582 & .719 & .820 & .352 & .269 & .377 & .511\\
\hline
TeRo$^{\diamond}$ & .562 & .468 & .621 & .732 & .586 & .469 & .668 & .795 & .245 & .154 & .264 & .420\\
TComplEx$^{\diamond}$ & .619 & .542 & .661 & .767 & .665 & .583 & .716 & .811 & .346 & .259 & .372 & .515\\
TeLM$^{\dagger}$ & .625 & .545 & .673 & .774 & .678 & .599 & .728 & .823 & .350 & .261 & .375 & .504\\
TLT-KGE(C)$^{\diamond}$ & .630 & .549 & .678 & .777 & .686 & .607 & .735 & .831 & .356 & .267 & .385 & .532\\
TeAST$^{\dagger}$ & \underline{.637} & \underline{.560} & .682 & .782 & .683 & .604 & .732 & .829 & \underline{.371} & \underline{.283} & \underline{.401} & \underline{.544}\\
\hline
RotatQVS$^{\diamond}$ & .591 & .507 & .642 & .754 & .633 & .529 & .709 & .813 & .270 & .175 & .293 & .458\\
TLT-KGE(Q)$^{\diamond}$ & .634 & .551 & \underline{.684} & \underline{.786} & \underline{.690} & \underline{.609} & \underline{.741} & \underline{.835} & .358 & .265 & .388 & .543\\
\hline
TQuatE & \bf.639 & \bf.561 & \bf.686 & \bf.787 & \bf.694 & \bf.611 & \bf.748 & \bf.845 & \bf.388 & \bf.302 & \bf.417 & \bf.554\\
\hline
\end{tabular}}
\caption{Experimental results on ICEWS14, ICEWS05-15 and GDELT. * : results are taken from~\cite{xu20tero}, $\diamond$ : results are taken from ~\cite{10.1145/3511808.3557233-TLTKGE}, $\dagger$ : results are taken from ~\cite{DBLP:conf/acl/LiSG23TeAST}. "-" means that results are not reported in those papers. The best score is in \textbf{bold} and second best score is \underline{underline}.}
\label{tab:results}
\end{table*}

\subsection{Settings}

\subsubsection{Evaluation Protocol}

The TKGC task, which focuses on predicting incomplete temporal facts with a missing entity $(h,r,?,\tau)$ or $(?,r,t,\tau)$, is employed to evaluate the performance of the proposed model.
For the inference stage, we adhere to strong baselines~\cite{10.1145/3511808.3557233-TLTKGE,DBLP:conf/acl/LiSG23TeAST}, where we substitute $h$ and $t$ with each entity from $\mathcal{E}$ for every quadruple in the test set. Subsequently, we compute scores for all corrupted quadruples (i.e., $(h,r,?,\tau)$ or $(?,r,t,\tau)$) and rank all candidate entities based on these scores, considering the time-wise filtered settings~\cite{10.1145/3511808.3557233-TLTKGE,DBLP:conf/acl/LiSG23TeAST}. 

The model's performance is evaluated using standard evaluation metrics, including Mean Reciprocal Rank (MRR) and Hits@$k$ (with $k=1,3,10$). MRR calculates the average reciprocal of the ranks, whereas Hits@$k$ measures the proportion of correct entities within the top $k$ predictions. Hits@1 specifically denotes the accuracy of predictions. Higher values of MRR and Hits@$k$ signify superior performance.

\subsubsection{Implementation Details}

We implemented our proposed model, TQuatE, using PyTorch on three datasets. The embedding dimension $d$ is set to $2000$ for all datasets. We optimized our model with Adagrad~\cite{DBLP:journals/jmlr/DuchiHS11-adagrad}, and the learning rate is set to 0.1. The maximum training epoch is set to 200. The norm $p$ is set to 4.

For hyperparameter tuning, we performed a grid search to find the best weights for the embedding regularizer ($\lambda_e$) and the periodic temporal regularizer ($\lambda_{\tau}$). The optimal hyperparameters for ICEWS14, ICEWS05-15 are
$\lambda_e = 0.0025, \ \lambda_{\tau} = 0.1$, and GDELT are $\lambda_e = 0.0001, \ \lambda_{\tau} = 0.1$
The training time for ICEWS14 is less than half an hour. For ICEWS05-15, it took about 2 hours, and for GDELT, it took about 6 hours. 
The reported performances are the averages of five independent runs.

\subsection{Main Results}

Table~\ref{tab:results} lists the main experimental results of our proposed model and all baselines on the datasets ICEWS14, ICEWS05-15, and GDELT. Both static and time-aware models modeled in hypercomplex space outperform those modeled in real and complex spaces. As time-aware models can capture the evolution of KGs over time, they achieve better performance in TKGC tasks compared to static models. Our proposed model TQuatE represents the KGs in hypercomplex space, offering more degrees of freedom, thus outperforming the SOTA method TeAST modeled in complex space. The improvements of MRR compared to TeAST on the three datasets are 0.31\%, 1.61\%, and 4.58\%, respectively. Due to the adoption of a more flexible periodic time representation, TQuatE also outperforms the SOTA model TLT-KGE(Q) modeled in quaternion space. Compared to TLT-KGE(Q), TQuatE improves the MRR on the three datasets by 0.79\%, 0.58\%, and 8.38\%, respectively. 
BoxTE~\cite{DBLP:conf/aaai/MessnerAC22BoxTE} emphasizes the need for a high level of temporal inductive capacity to capture the significant temporal variability in GDELT effectively. Our proposed model significantly outperforms the existing SOTA model on GDELT, suggesting that both the quaternion representation and periodic time enhance the capability to model complex temporal variability.

\subsection{Ablation Study}

We conduct an ablation experiment on ICEWS14 to investigate the contributions of modeling periodic time.

\begin{table}
\centering
\resizebox{0.9\linewidth}{!}{
\begin{tabular}{ccccc}
\hline
& {MRR} & {Hits@1} & {Hits@3} & {Hits@10} \\
\hline
TQuatE & \bf.639 & \bf.561 & \bf.686 & \bf.787 \\
 w/o PT & \bf.629 & \bf.551 & \bf.675 & \bf.774 \\
\hline
\end{tabular}}
\caption{Ablation study of TQuatE on ICEWS14. w/o means without.}
\label{tab:ablation}
\end{table}

As reported in Table~\ref{tab:ablation}, without periodic time (w/o PT), all metrics of the performance of TQuatE drops, which indicates the effectiveness of model the periodic time.

\subsection{HyperParameter Analysis}

\begin{figure}
  \centering  
  \includegraphics[width=1.0\linewidth]{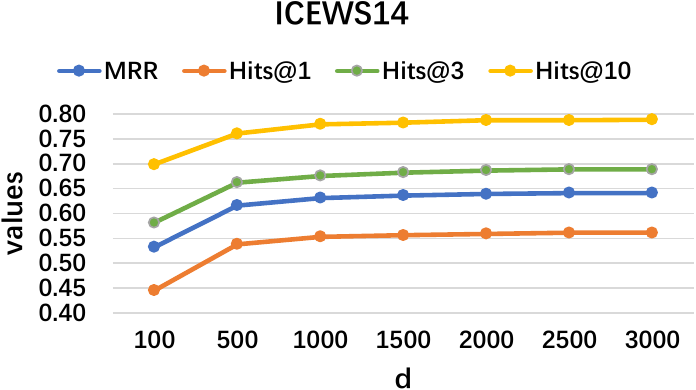}  
  \caption{TKGC performance for different embedding dimensions $d$ on ICEWS14.}
  \label{fig:dim}
\end{figure}

We conduct experiments on the following hyper-parameters to investigate their effect on the performance of TQuatE. 

(1) The embedding dimension $d$. 

We choose the dimension from the set $\{100, 500, 1000, 1500, 2000, 2500, 3000\}$ and conduct the experiments on the ICEWS14 dataset. 

From the experimental results depicted in Figure~\ref{fig:dim}, we note a consistent enhancement in the performance of TQuatE as the embedding dimension increases, reaching an optimal level at $2000$. However, the rate of performance improvement gradually diminishes for dimensions larger than $2000$. Therefore, we opt for an embedding dimension of $2000$ as it balances achieving optimal performance with minimizing time and space costs on the ICEWS14 dataset.

(2) The weight of embedding regularizer $\lambda_e$ and time regularizer $\lambda_\tau$.

\begin{figure}
  \centering  
  \includegraphics[width=0.99\linewidth]{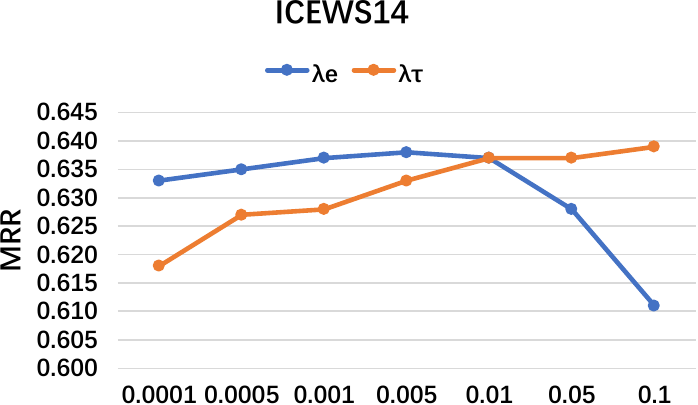}  
  \caption{MRR values for different weight of embedding regularizer $\lambda_e$ and time regularizer $\lambda_\tau$ on ICEWS14.}
  \label{fig:lambda}
\end{figure}

\begin{table}
\centering
\resizebox{\linewidth}{!}{
\begin{tabular}{l l}
\hline
{Model} & {Number of Parameters}\\
\hline
{TComplEx} & {$(|\mathcal{E}| + |\mathcal{R}| \times 2 +|\mathcal{T}|) \times d \times 2$}\\
{Tero} & {$(|\mathcal{E}| + |\mathcal{R}| \times 2) \times d \times 2 + |\mathcal{T}| \times d$}\\
{TeLM} & {$(|\mathcal{E}| + |\mathcal{R}| \times 2 +|\mathcal{T}|) \times d \times 2 + d$}\\
{TLT-KGE(C)} & {$(|\mathcal{E}| + |\mathcal{R}|\times 2 + |\mathcal{T}|\times 3 + m) \times d \times 2$}\\
{TeAST} & {$(|\mathcal{E}| + |\mathcal{R}| \times 2 +|\mathcal{T}| \times 2) \times d \times 2$}\\
\hline
{RotateQVS} & {$(|\mathcal{E}| + |\mathcal{R}| \times 2 +|\mathcal{T}|) \times d \times 4$}\\
{TLT-KGE(Q)} & {$(|\mathcal{E}| + |\mathcal{R}|\times 2 + |\mathcal{T}|\times 3 + m) \times d \times 2$}\\
\hline
{TQuatE} & {$(|\mathcal{E}| + |\mathcal{R}|\times 2 + |\mathcal{T}|\times 2) \times d \times 2$}\\
\hline
\end{tabular}}
\caption{The number of parameters of the models. $|\mathcal{E}|$, $|\mathcal{R}|$, $|\mathcal{T}|$, $d$, and $m$ denote the number of entities, relations, timestamps, embedding dimension, and time windows, respectively.}
\label{tab:complexity}
\end{table}

We select the weight of $\lambda_e$ and $\lambda_\tau$ from the set $\{0.0001, 0.0005, 0.001, 0.005, 0.01, 0.05, 0.1\}$ and perform experiments on the ICEWS14 dataset. 

From Figure~\ref{fig:lambda}, we can see that the MRR reaches its optimum when $\lambda_e$ is around 0.005, while for $\lambda_\tau$, the optimum MRR is achieved at around 0.1. This suggests that the impact of temporal regularization during training is greater than that of embedding regularization.

\subsection{Complexity Analysis}


We present the number of parameters during the training phase of competitive TKGC models in Table~\ref{tab:complexity}. TComplEx, Tero, TeLM, TLT-KGE(C), and TeAST work within the complex space, while RotateQVS, TLT-KGE(Q), and our proposed model TQuatE operate within the quaternion space.

From the table, we observe that TQuatE and TeAST have equal parameters since they share the same embedding dimension ($d=2000$). However, owing to the increased degrees of freedom in quaternion space compared to complex space, TQuatE outperforms TeAST. 
Despite TLT-KGE utilizing dimensions of 1200 in ICEWS and 1500 in GDELT, respectively, the number of parameters is comparable to our proposed method due to the inclusion of additional temporal modeling requirements. TQuatE outperforms TLT-KGE, highlighting the effectiveness of modeling periodic time.


\section{Conclusion}

In conclusion, this paper proposes a novel approach for TKGC by leveraging quaternion embeddings in hypercomplex space. Our method effectively models the dynamic evolution of facts over time, incorporating both temporal and periodic features of relations. Through extensive experiments on multiple datasets, including ICEWS14, ICEWS05-15, and GDELT, we demonstrate that our model achieves SOTA performance, particularly excelling in datasets with complex temporal variability like GDELT. 

\section{Limitations}
Although TQuatE employs the same parameters as TeAST, the quaternion multiplication operation is more complex than complex multiplication. Therefore, our method requires more time than TeAST. However, the performance improvement on the ICEWS14 and ICEWS05-15 datasets is insignificant. Table~\ref{tab:limitation} shows the MRR values and time cost of TQuatE and TeAST on the ICEWS14 dataset.

\begin{table}[h]
\centering
\resizebox{0.7\linewidth}{!}{
\begin{tabular}{lcc}
\hline
 & {MRR} & {Time Cost (s)} \\
\hline
TeAST & 0.637 &  1487 \\
TQuatE & 0.639 & 2011 \\
\hline
\end{tabular}}
\caption{MRR values and time cost of TQuatE and TeAST on the ICEWS14.}
\label{tab:limitation}
\end{table}

\bibliography{custom}

\begin{thebibliography}{27}
\expandafter\ifx\csname natexlab\endcsname\relax\def\natexlab#1{#1}\fi

\bibitem[{Abboud et~al.(2020)Abboud, Ceylan, Lukasiewicz, and Salvatori}]{DBLP:conf/nips/AbboudCLS20boxe}
Ralph Abboud, {\.I}smail~{\.I}lkan Ceylan, Thomas Lukasiewicz, and Tommaso Salvatori. 2020.
\newblock \href {https://proceedings.neurips.cc/paper/2020/hash/6dbbe6abe5f14af882ff977fc3f35501-Abstract.html} {Boxe: {A} box embedding model for knowledge base completion}.
\newblock In \emph{Advances in Neural Information Processing Systems 33: Annual Conference on Neural Information Processing Systems 2020, NeurIPS 2020, December 6-12, 2020, virtual}.

\bibitem[{Aggarwal et~al.(2001)Aggarwal, Hinneburg, and Keim}]{DBLP:conf/icdt/AggarwalHK01-distance1}
Charu~C. Aggarwal, Alexander Hinneburg, and Daniel~A. Keim. 2001.
\newblock \href {https://doi.org/10.1007/3-540-44503-X\_27} {On the surprising behavior of distance metrics in high dimensional spaces}.
\newblock In \emph{Database Theory - {ICDT} 2001, 8th International Conference, London, UK, January 4-6, 2001, Proceedings}, volume 1973 of \emph{Lecture Notes in Computer Science}, pages 420--434. Springer.

\bibitem[{Bordes et~al.(2013)Bordes, Usunier, Garc{\'{\i}}a{-}Dur{\'{a}}n, Weston, and Yakhnenko}]{DBLP:conf/nips/BordesUGWY13TransE}
Antoine Bordes, Nicolas Usunier, Alberto Garc{\'{\i}}a{-}Dur{\'{a}}n, Jason Weston, and Oksana Yakhnenko. 2013.
\newblock \href {https://proceedings.neurips.cc/paper/2013/hash/1cecc7a77928ca8133fa24680a88d2f9-Abstract.html} {Translating embeddings for modeling multi-relational data}.
\newblock In \emph{Advances in Neural Information Processing Systems 26: 27th Annual Conference on Neural Information Processing Systems 2013. Proceedings of a meeting held December 5-8, 2013, Lake Tahoe, Nevada, United States}, pages 2787--2795.

\bibitem[{Chen et~al.(2022)Chen, Wang, Li, and Li}]{chen-etal-2022-rotateqvs}
Kai Chen, Ye~Wang, Yitong Li, and Aiping Li. 2022.
\newblock \href {https://doi.org/10.18653/v1/2022.acl-long.402} {{R}otate{QVS}: Representing temporal information as rotations in quaternion vector space for temporal knowledge graph completion}.
\newblock In \emph{Proceedings of the 60th Annual Meeting of the Association for Computational Linguistics (Volume 1: Long Papers)}, pages 5843--5857, Stroudsburg, PA, USA. Association for Computational Linguistics.

\bibitem[{Chen et~al.(2023)Chen, Guo, Li, and Li}]{DBLP:conf/sigir/ChenGL023RS}
Qian Chen, Zhiqiang Guo, Jianjun Li, and Guohui Li. 2023.
\newblock \href {https://doi.org/10.1145/3539618.3591706} {Knowledge-enhanced multi-view graph neural networks for session-based recommendation}.
\newblock In \emph{Proceedings of the 46th International {ACM} {SIGIR} Conference on Research and Development in Information Retrieval, {SIGIR} 2023, Taipei, Taiwan, July 23-27, 2023}, pages 352--361. {ACM}.

\bibitem[{Dasgupta et~al.(2018)Dasgupta, Ray, and Talukdar}]{dasgupta18hyte}
Shib~Sankar Dasgupta, Swayambhu~Nath Ray, and Partha Talukdar. 2018.
\newblock \href {https://doi.org/10.18653/v1/D18-1225} {{H}y{TE}: Hyperplane-based temporally aware knowledge graph embedding}.
\newblock In \emph{Proceedings of the 2018 Conference on Empirical Methods in Natural Language Processing}, EMNLP ’18, pages 2001--2011, Stroudsburg, PA, USA. Association for Computational Linguistics.

\bibitem[{Duchi et~al.(2011)Duchi, Hazan, and Singer}]{DBLP:journals/jmlr/DuchiHS11-adagrad}
John~C. Duchi, Elad Hazan, and Yoram Singer. 2011.
\newblock \href {https://doi.org/10.5555/1953048.2021068} {Adaptive subgradient methods for online learning and stochastic optimization}.
\newblock \emph{J. Mach. Learn. Res.}, 12:2121--2159.

\bibitem[{Garc{\'{\i}}a{-}Dur{\'{a}}n et~al.(2018)Garc{\'{\i}}a{-}Dur{\'{a}}n, Dumancic, and Niepert}]{Garcia-DuranDN18tatranse}
Alberto Garc{\'{\i}}a{-}Dur{\'{a}}n, Sebastijan Dumancic, and Mathias Niepert. 2018.
\newblock \href {https://doi.org/10.18653/v1/D18-1516} {Learning sequence encoders for temporal knowledge graph completion}.
\newblock In \emph{Proceedings of the 2018 Conference on Empirical Methods in Natural Language Processing}, EMNLP ’18, pages 4816--4821, Stroudsburg, PA, USA. Association for Computational Linguistics.

\bibitem[{Hamilton(1843)}]{hamilton1843quaternions}
William~Rowan Hamilton. 1843.
\newblock On quaternions; or on a new system of imaginaries in algebra (letter to john t. graves, dated october 17, 1843).
\newblock \emph{Philos. Magazine}, 25:489--495.

\bibitem[{Lacroix et~al.(2020)Lacroix, Obozinski, and Usunier}]{LacroixOU20TComplEx}
Timoth{\'{e}}e Lacroix, Guillaume Obozinski, and Nicolas Usunier. 2020.
\newblock \href {https://openreview.net/forum?id=rke2P1BFwS} {Tensor decompositions for temporal knowledge base completion}.
\newblock In \emph{8th International Conference on Learning Representations, {ICLR} 2020}, Addis Ababa, Ethiopia. OpenReview.net.

\bibitem[{Lautenschlager et~al.(2015)Lautenschlager, Shellman, and Ward}]{DVN/28117_2015-icews}
Jennifer Lautenschlager, Steve Shellman, and Michael Ward. 2015.
\newblock \href {https://doi.org/10.7910/DVN/28117} {{ICEWS Event Aggregations}}.

\bibitem[{Leblay and Chekol(2018)}]{10.1145/3184558.3191639-TTransE}
Julien Leblay and Melisachew~Wudage Chekol. 2018.
\newblock \href {https://doi.org/10.1145/3184558.3191639} {Deriving validity time in knowledge graph}.
\newblock In \emph{Companion Proceedings of the The Web Conference 2018}, WWW '18, page 1771–1776, Republic and Canton of Geneva, CHE. International World Wide Web Conferences Steering Committee.

\bibitem[{Leetaru and Schrodt(2013)}]{leetaru2013gdelt}
Kalev Leetaru and Philip~A Schrodt. 2013.
\newblock \href {http://data.gdeltproject.org/documentation/ISA.2013.GDELT.pdf} {Gdelt: Global data on events, location, and tone, 1979--2012}.
\newblock In \emph{ISA annual convention}, volume~2, pages 1--49, State College, PA, USA. Citeseer, Citeseer.

\bibitem[{Li et~al.(2023)Li, Su, and Gao}]{DBLP:conf/acl/LiSG23TeAST}
Jiang Li, Xiangdong Su, and Guanglai Gao. 2023.
\newblock \href {https://doi.org/10.18653/V1/2023.ACL-LONG.862} {Teast: Temporal knowledge graph embedding via archimedean spiral timeline}.
\newblock In \emph{Proceedings of the 61st Annual Meeting of the Association for Computational Linguistics (Volume 1: Long Papers), {ACL} 2023, Toronto, Canada, July 9-14, 2023}, pages 15460--15474. Association for Computational Linguistics.

\bibitem[{Liang et~al.(2023)Liang, Meng, Liu, Liu, Tu, Wang, Zhou, and Liu}]{DBLP:conf/sigir/0006MLLTWZL23rpc}
Ke~Liang, Lingyuan Meng, Meng Liu, Yue Liu, Wenxuan Tu, Siwei Wang, Sihang Zhou, and Xinwang Liu. 2023.
\newblock \href {https://doi.org/10.1145/3539618.3591711} {Learn from relational correlations and periodic events for temporal knowledge graph reasoning}.
\newblock In \emph{Proceedings of the 46th International {ACM} {SIGIR} Conference on Research and Development in Information Retrieval, {SIGIR} 2023, Taipei, Taiwan, July 23-27, 2023}, pages 1559--1568. {ACM}.

\bibitem[{Messner et~al.(2022)Messner, Abboud, and Ceylan}]{DBLP:conf/aaai/MessnerAC22BoxTE}
Johannes Messner, Ralph Abboud, and {\.I}smail~{\.I}lkan Ceylan. 2022.
\newblock \href {https://doi.org/10.1609/AAAI.V36I7.20746} {Temporal knowledge graph completion using box embeddings}.
\newblock In \emph{Thirty-Sixth {AAAI} Conference on Artificial Intelligence, {AAAI} 2022, Thirty-Fourth Conference on Innovative Applications of Artificial Intelligence, {IAAI} 2022, The Twelveth Symposium on Educational Advances in Artificial Intelligence, {EAAI} 2022 Virtual Event, February 22 - March 1, 2022}, pages 7779--7787. {AAAI} Press.

\bibitem[{Nguyen et~al.(2022)Nguyen, Vu, Nguyen, and Phung}]{DBLP:conf/www/NguyenVNP22QuatRE}
Dai~Quoc Nguyen, Thanh Vu, Tu~Dinh Nguyen, and Dinh~Q. Phung. 2022.
\newblock \href {https://doi.org/10.1145/3487553.3524251} {Quatre: Relation-aware quaternions for knowledge graph embeddings}.
\newblock In \emph{Companion of The Web Conference 2022, Virtual Event / Lyon, France, April 25 - 29, 2022}, pages 189--192. {ACM}.

\bibitem[{Sadeghian et~al.(2021)Sadeghian, Armandpour, Colas, and Wang}]{sadeghian2021chronor}
Ali Sadeghian, Mohammadreza Armandpour, Anthony Colas, and Daisy~Zhe Wang. 2021.
\newblock \href {https://doi.org/10.1609/aaai.v35i7.16802} {Chronor: Rotation based temporal knowledge graph embedding}.
\newblock In \emph{Proceedings of the AAAI Conference on Artificial Intelligence, {AAAI}2021}, volume~35, pages 6471--6479, Palo Alto, CA, USA. {AAAI} Press.

\bibitem[{Sun et~al.(2019)Sun, Deng, Nie, and Tang}]{DBLP:conf/iclr/SunDNT19-rotate}
Zhiqing Sun, Zhi{-}Hong Deng, Jian{-}Yun Nie, and Jian Tang. 2019.
\newblock \href {https://openreview.net/forum?id=HkgEQnRqYQ} {Rotate: Knowledge graph embedding by relational rotation in complex space}.
\newblock In \emph{7th International Conference on Learning Representations, {ICLR} 2019}, New Orleans, LA, USA. OpenReview.net.

\bibitem[{Trouillon et~al.(2016)Trouillon, Welbl, Riedel, Gaussier, and Bouchard}]{DBLP:conf/icml/TrouillonWRGB16ComplEx}
Th{\'{e}}o Trouillon, Johannes Welbl, Sebastian Riedel, {\'{E}}ric Gaussier, and Guillaume Bouchard. 2016.
\newblock \href {http://proceedings.mlr.press/v48/trouillon16.html} {Complex embeddings for simple link prediction}.
\newblock In \emph{Proceedings of the 33nd International Conference on Machine Learning, {ICML} 2016, New York City, NY, USA, June 19-24, 2016}, volume~48 of \emph{{JMLR} Workshop and Conference Proceedings}, pages 2071--2080. JMLR.org.

\bibitem[{Wang et~al.(2023)Wang, Zhang, Liang, and Li}]{DBLP:conf/acl/WangZLL23QA}
Yujie Wang, Hu~Zhang, Jiye Liang, and Ru~Li. 2023.
\newblock \href {https://doi.org/10.18653/V1/2023.ACL-LONG.785} {Dynamic heterogeneous-graph reasoning with language models and knowledge representation learning for commonsense question answering}.
\newblock In \emph{Proceedings of the 61st Annual Meeting of the Association for Computational Linguistics (Volume 1: Long Papers), {ACL} 2023, Toronto, Canada, July 9-14, 2023}, pages 14048--14063. Association for Computational Linguistics.

\bibitem[{Xu et~al.(2021)Xu, Chen, Nayyeri, and Lehmann}]{DBLP:conf/naacl/XuCNL21TeLM}
Chengjin Xu, Yung{-}Yu Chen, Mojtaba Nayyeri, and Jens Lehmann. 2021.
\newblock \href {https://doi.org/10.18653/V1/2021.NAACL-MAIN.202} {Temporal knowledge graph completion using a linear temporal regularizer and multivector embeddings}.
\newblock In \emph{Proceedings of the 2021 Conference of the North American Chapter of the Association for Computational Linguistics: Human Language Technologies, {NAACL-HLT} 2021, Online, June 6-11, 2021}, pages 2569--2578. Association for Computational Linguistics.

\bibitem[{Xu et~al.(2020)Xu, Nayyeri, Alkhoury, Shariat~Yazdi, and Lehmann}]{xu20tero}
Chengjin Xu, Mojtaba Nayyeri, Fouad Alkhoury, Hamed Shariat~Yazdi, and Jens Lehmann. 2020.
\newblock \href {https://doi.org/10.18653/v1/2020.coling-main.139} {{T}e{R}o: A time-aware knowledge graph embedding via temporal rotation}.
\newblock In \emph{Proceedings of the 28th International Conference on Computational Linguistics}, pages 1583--1593, Stroudsburg, PA, USA. Association for Computational Linguistics.

\bibitem[{Zhang et~al.(2022)Zhang, Zhang, Ao, Zhuang, Xu, and He}]{10.1145/3511808.3557233-TLTKGE}
Fuwei Zhang, Zhao Zhang, Xiang Ao, Fuzhen Zhuang, Yongjun Xu, and Qing He. 2022.
\newblock \href {https://doi.org/10.1145/3511808.3557233} {Along the time: Timeline-traced embedding for temporal knowledge graph completion}.
\newblock In \emph{Proceedings of the 31st ACM International Conference on Information \& Knowledge Management}, CIKM '22, page 2529–2538, New York, NY, USA. Association for Computing Machinery.

\bibitem[{Zhang et~al.(2019)Zhang, Tay, Yao, and Liu}]{DBLP:conf/nips/0007TYL19QuatE}
Shuai Zhang, Yi~Tay, Lina Yao, and Qi~Liu. 2019.
\newblock \href {https://proceedings.neurips.cc/paper/2019/hash/d961e9f236177d65d21100592edb0769-Abstract.html} {Quaternion knowledge graph embeddings}.
\newblock In \emph{Advances in Neural Information Processing Systems 32: Annual Conference on Neural Information Processing Systems 2019, NeurIPS 2019, December 8-14, 2019, Vancouver, BC, Canada}, pages 2731--2741.

\bibitem[{Zhao et~al.(2023)Zhao, Chen, Xing, and Miao}]{DBLP:journals/tnn/ZhaoCXM23SE}
Xuejiao Zhao, Huanhuan Chen, Zhenchang Xing, and Chunyan Miao. 2023.
\newblock \href {https://doi.org/10.1109/TNNLS.2021.3113026} {Brain-inspired search engine assistant based on knowledge graph}.
\newblock \emph{{IEEE} Trans. Neural Networks Learn. Syst.}, 34(8):4386--4400.

\bibitem[{Zimek et~al.(2012)Zimek, Schubert, and Kriegel}]{DBLP:journals/sadm/ZimekSK12-distance2}
Arthur Zimek, Erich Schubert, and Hans{-}Peter Kriegel. 2012.
\newblock \href {https://doi.org/10.1002/SAM.11161} {A survey on unsupervised outlier detection in high-dimensional numerical data}.
\newblock \emph{Stat. Anal. Data Min.}, 5(5):363--387.

\end{thebibliography}

\appendix

\section{Proof of Propositions 1}
\label{sec:appendix A}
To demonstrate that TQuatE can model symmetric relation pattern, that is $\forall h, t, \tau, r(h, t, \tau,) \land r(t, h, \tau)$ hold True accordding Definetion~\ref{def1}, which implies $\mathbf{q}_{h} \otimes \mathbf{q'}_{r(\tau)} \cdot \mathbf{q}_t = \mathbf{q}_{t} \otimes \mathbf{q'}_{r(\tau)} \cdot \mathbf{q}_h$. The proof is as follows:
\begin{equation}
\resizebox{1.0\linewidth}{!}{
$
\begin{aligned}
   & \mathbf{q}_{h} \otimes \mathbf{q'}_{r(\tau)} \cdot \mathbf{q}_t \\   
   = & \ (\mathbf{e}^{a}_{h} + \mathbf{e}^{b}_{h}\mathbf{i} + \mathbf{e}^{c}_{h}\mathbf{j} + \mathbf{e}^{d}_{h}\mathbf{k}) \ \otimes \\   
   & \ (\mathbf{e}^{a}_{r(\tau)} + \mathbf{e}^{b}_{r(\tau)}\mathbf{i} + \mathbf{e}^{c}_{r(\tau)}\mathbf{j} + \mathbf{e}^{d}_{r(\tau)}\mathbf{k}) \ \cdot \\   
   & \ (\mathbf{e}^{a}_{t} + \mathbf{e}^{b}_{t}\mathbf{i} + \mathbf{e}^{c}_{t}\mathbf{j} + \mathbf{e}^{d}_{t}\mathbf{k}) \\ 
   = & \ [(\mathbf{e}^{a}_{h}\mathbf{e}^{a}_{r(\tau)} - \mathbf{e}^{b}_{h}\mathbf{e}^{b}_{r(\tau)} - \mathbf{e}^{c}_{h}\mathbf{e}^{c}_{r(\tau)} - \mathbf{e}^{d}_{h}\mathbf{e}^{d}_{r(\tau)}) + \\
   & \ (\mathbf{e}^{a}_{h}\mathbf{e}^{b}_{r(\tau)} + \mathbf{e}^{b}_{h}\mathbf{e}^{a}_{r(\tau)} + \mathbf{e}^{c}_{h}\mathbf{e}^{d}_{r(\tau)} - \mathbf{e}^{d}_{h}\mathbf{e}^{c}_{r(\tau)})\mathbf{i} + \\
   & \ (\mathbf{e}^{a}_{h}\mathbf{e}^{c}_{r(\tau)} - \mathbf{e}^{b}_{h}\mathbf{e}^{d}_{r(\tau)} + \mathbf{e}^{c}_{h}\mathbf{e}^{a}_{r(\tau)} + \mathbf{e}^{d}_{h}\mathbf{e}^{b}_{r(\tau)})\mathbf{j} + \\
   & \ (\mathbf{e}^{a}_{h}\mathbf{e}^{d}_{r(\tau)} + \mathbf{e}^{b}_{h}\mathbf{e}^{c}_{r(\tau)} - \mathbf{e}^{c}_{h}\mathbf{e}^{b}_{r(\tau)} + \mathbf{e}^{d}_{h}\mathbf{e}^{a}_{r(\tau)})\mathbf{k}]  \ \cdot \\
   & \ (\mathbf{e}^{a}_{t} + \mathbf{e}^{b}_{t}\mathbf{i} + \mathbf{e}^{c}_{t}\mathbf{j} + \mathbf{e}^{d}_{t}\mathbf{k}) \\ 
   = & \ (\mathbf{e}^{a}_{h}\mathbf{e}^{a}_{r(\tau)}\mathbf{e}^{a}_{t} - \mathbf{e}^{b}_{h}\mathbf{e}^{b}_{r(\tau)}\mathbf{e}^{a}_{t} - \mathbf{e}^{c}_{h}\mathbf{e}^{c}_{r(\tau)}\mathbf{e}^{a}_{t} - \mathbf{e}^{d}_{h}\mathbf{e}^{d}_{r(\tau)}\mathbf{e}^{a}_{t}) + \\
    & \ (\mathbf{e}^{a}_{h}\mathbf{e}^{b}_{r(\tau)}\mathbf{e}^{b}_{t} + \mathbf{e}^{b}_{h}\mathbf{e}^{a}_{r(\tau)}\mathbf{e}^{b}_{t} + \mathbf{e}^{c}_{h}\mathbf{e}^{d}_{r(\tau)}\mathbf{e}^{b}_{t} - \mathbf{e}^{d}_{h}\mathbf{e}^{c}_{r(\tau)}\mathbf{e}^{b}_{t}) + \\
   & \ (\mathbf{e}^{a}_{h}\mathbf{e}^{c}_{r(\tau)}\mathbf{e}^{c}_{t} - \mathbf{e}^{b}_{h}\mathbf{e}^{d}_{r(\tau)}\mathbf{e}^{c}_{t} + \mathbf{e}^{c}_{h}\mathbf{e}^{a}_{r(\tau)}\mathbf{e}^{c}_{t} + \mathbf{e}^{d}_{h}\mathbf{e}^{b}_{r(\tau)}\mathbf{e}^{c}_{t}) + \\
   & \ (\mathbf{e}^{a}_{h}\mathbf{e}^{d}_{r(\tau)}\mathbf{e}^{d}_{t} + \mathbf{e}^{b}_{h}\mathbf{e}^{c}_{r(\tau)}\mathbf{e}^{d}_{t} - \mathbf{e}^{c}_{h}\mathbf{e}^{b}_{r(\tau)}\mathbf{e}^{d}_{t} + \mathbf{e}^{d}_{h}\mathbf{e}^{a}_{r(\tau)}\mathbf{e}^{d}_{t}) \\ 
   \label{eq12}
\end{aligned}
$}
\end{equation}

\begin{equation}
\resizebox{1.0\linewidth}{!}{
$
\begin{aligned}
   & \mathbf{q}_{t} \otimes \mathbf{q'}_{r(\tau)} \cdot \mathbf{q}_h \\   
   = & \ (\mathbf{e}^{a}_{t} + \mathbf{e}^{b}_{t}\mathbf{i} + \mathbf{e}^{c}_{t}\mathbf{j} + \mathbf{e}^{d}_{t}\mathbf{k}) \ \otimes \\   
   & \ (\mathbf{e}^{a}_{r(\tau)} + \mathbf{e}^{b}_{r(\tau)}\mathbf{i} + \mathbf{e}^{c}_{r(\tau)}\mathbf{j} + \mathbf{e}^{d}_{r(\tau)}\mathbf{k}) \ \cdot \\   
   & \ (\mathbf{e}^{a}_{h} + \mathbf{e}^{b}_{h}\mathbf{i} + \mathbf{e}^{c}_{h}\mathbf{j} + \mathbf{e}^{d}_{h}\mathbf{k}) \\ 
   = & \ [(\mathbf{e}^{a}_{t}\mathbf{e}^{a}_{r(\tau)} - \mathbf{e}^{b}_{t}\mathbf{e}^{b}_{r(\tau)} - \mathbf{e}^{c}_{t}\mathbf{e}^{c}_{r(\tau)} - \mathbf{e}^{d}_{t}\mathbf{e}^{d}_{r(\tau)}) + \\
   & \ (\mathbf{e}^{a}_{t}\mathbf{e}^{b}_{r(\tau)} + \mathbf{e}^{b}_{t}\mathbf{e}^{a}_{r(\tau)} + \mathbf{e}^{c}_{t}\mathbf{e}^{d}_{r(\tau)} - \mathbf{e}^{d}_{t}\mathbf{e}^{c}_{r(\tau)})\mathbf{i} + \\
   & \ (\mathbf{e}^{a}_{t}\mathbf{e}^{c}_{r(\tau)} - \mathbf{e}^{b}_{t}\mathbf{e}^{d}_{r(\tau)} + \mathbf{e}^{c}_{t}\mathbf{e}^{a}_{r(\tau)} + \mathbf{e}^{d}_{t}\mathbf{e}^{b}_{r(\tau)})\mathbf{j} + \\
   & \ (\mathbf{e}^{a}_{t}\mathbf{e}^{d}_{r(\tau)} + \mathbf{e}^{b}_{t}\mathbf{e}^{c}_{r(\tau)} - \mathbf{e}^{c}_{t}\mathbf{e}^{b}_{r(\tau)} + \mathbf{e}^{d}_{t}\mathbf{e}^{a}_{r(\tau)})\mathbf{k}]  \ \cdot \\
   & \ (\mathbf{e}^{a}_{h} + \mathbf{e}^{b}_{h}\mathbf{i} + \mathbf{e}^{c}_{h}\mathbf{j} + \mathbf{e}^{d}_{h}\mathbf{k}) \\ 
   = & \ (\mathbf{e}^{a}_{t}\mathbf{e}^{a}_{r(\tau)}\mathbf{e}^{a}_{h} - \mathbf{e}^{b}_{t}\mathbf{e}^{b}_{r(\tau)}\mathbf{e}^{a}_{h} - \mathbf{e}^{c}_{t}\mathbf{e}^{c}_{r(\tau)}\mathbf{e}^{a}_{h} - \mathbf{e}^{d}_{t}\mathbf{e}^{d}_{r(\tau)}\mathbf{e}^{a}_{h}) + \\
    & \ (\mathbf{e}^{a}_{t}\mathbf{e}^{b}_{r(\tau)}\mathbf{e}^{b}_{h} + \mathbf{e}^{b}_{t}\mathbf{e}^{a}_{r(\tau)}\mathbf{e}^{b}_{h} + \mathbf{e}^{c}_{t}\mathbf{e}^{d}_{r(\tau)}\mathbf{e}^{b}_{h} - \mathbf{e}^{d}_{t}\mathbf{e}^{c}_{r(\tau)}\mathbf{e}^{b}_{h}) + \\
   & \ (\mathbf{e}^{a}_{t}\mathbf{e}^{c}_{r(\tau)}\mathbf{e}^{c}_{h} - \mathbf{e}^{b}_{t}\mathbf{e}^{d}_{r(\tau)}\mathbf{e}^{c}_{h} + \mathbf{e}^{c}_{t}\mathbf{e}^{a}_{r(\tau)}\mathbf{e}^{c}_{h} + \mathbf{e}^{d}_{t}\mathbf{e}^{b}_{r(\tau)}\mathbf{e}^{c}_{h}) + \\
   & \ (\mathbf{e}^{a}_{t}\mathbf{e}^{d}_{r(\tau)}\mathbf{e}^{d}_{h} + \mathbf{e}^{b}_{t}\mathbf{e}^{c}_{r(\tau)}\mathbf{e}^{d}_{h} - \mathbf{e}^{c}_{t}\mathbf{e}^{b}_{r(\tau)}\mathbf{e}^{d}_{h} + \mathbf{e}^{d}_{t}\mathbf{e}^{a}_{r(\tau)}\mathbf{e}^{d}_{h}) \\ 
   \label{eq13}
\end{aligned}
$}
\end{equation}

if the imaginary part of $\mathbf{q'}_{r(\tau)}$ is zero, i.e. $\mathbf{q'}_{r(\tau)} = \mathbf{e}^{a}_{r(\tau)}$, then:

\begin{equation}
\resizebox{1.0\linewidth}{!}{
$
\begin{aligned}
   & \mathbf{q}_{h} \otimes \mathbf{q'}_{r(\tau)} \cdot \mathbf{q}_t \\   
   = & \ \mathbf{e}^{a}_{h}\mathbf{e}^{a}_{r(\tau)}\mathbf{e}^{a}_{t} + \mathbf{e}^{b}_{h}\mathbf{e}^{a}_{r(\tau)}\mathbf{e}^{b}_{t} - \mathbf{e}^{c}_{h}\mathbf{e}^{a}_{r(\tau)}\mathbf{e}^{c}_{t} + \mathbf{e}^{d}_{h}\mathbf{e}^{a}_{r(\tau)}\mathbf{e}^{d}_{t} \\ 
   \label{eq14}
\end{aligned}
$}
\end{equation}

\begin{equation}
\resizebox{1.0\linewidth}{!}{
$
\begin{aligned}
   & \mathbf{q}_{t} \otimes \mathbf{q'}_{r(\tau)} \cdot \mathbf{q}_h \\   
   = & \ \mathbf{e}^{a}_{t}\mathbf{e}^{a}_{r(\tau)}\mathbf{e}^{a}_{h} + \mathbf{e}^{b}_{t}\mathbf{e}^{a}_{r(\tau)}\mathbf{e}^{b}_{h} - \mathbf{e}^{c}_{t}\mathbf{e}^{a}_{r(\tau)}\mathbf{e}^{c}_{h} + \mathbf{e}^{d}_{t}\mathbf{e}^{a}_{r(\tau)}\mathbf{e}^{d}_{h} \\ 
   \label{eq15}
\end{aligned}
$}
\end{equation}

By comparison, the above two equations are identical. Therefore, when the imaginary part of $\mathbf{q'}_{r(\tau)}$ is zero, TQuatE can model symmetric relations.

\section{Proof of Propositions 2}
\label{sec:appendix B}
To model the asymmetric relation pattern, that is 
 $\forall h, t, \tau, r(h, t, \tau,) \land \lnot r(t, h, \tau)$ hold True accordding Definetion~\ref{def2}, which implies $\mathbf{q}_{h} \otimes \mathbf{q'}_{r(\tau)} \cdot \mathbf{q}_t \ \neq \mathbf{q}_{t} \otimes \mathbf{q'}_{r(\tau)} \cdot \mathbf{q}_h$. If the real part of $\mathbf{q'}_{r(\tau)}$ is zero, i.e. $\mathbf{q'}_{r(\tau)} = \mathbf{e}^{b}_{t}\mathbf{i} + \mathbf{e}^{c}_{t}\mathbf{j} + \mathbf{e}^{d}_{t}\mathbf{k}$, then:
\begin{equation}
\resizebox{1.0\linewidth}{!}{
$
\begin{aligned}
   & \mathbf{q}_{h} \otimes \mathbf{q'}_{r(\tau)} \cdot \mathbf{q}_t \\   
   = & \ ( - \mathbf{e}^{b}_{h}\mathbf{e}^{b}_{r(\tau)}\mathbf{e}^{a}_{t} - \mathbf{e}^{c}_{h}\mathbf{e}^{c}_{r(\tau)}\mathbf{e}^{a}_{t} - \mathbf{e}^{d}_{h}\mathbf{e}^{d}_{r(\tau)}\mathbf{e}^{a}_{t}) + \\
    & \ (\mathbf{e}^{a}_{h}\mathbf{e}^{b}_{r(\tau)}\mathbf{e}^{b}_{t} +  \mathbf{e}^{c}_{h}\mathbf{e}^{d}_{r(\tau)}\mathbf{e}^{b}_{t} - \mathbf{e}^{d}_{h}\mathbf{e}^{c}_{r(\tau)}\mathbf{e}^{b}_{t}) + \\
   & \ (\mathbf{e}^{a}_{h}\mathbf{e}^{c}_{r(\tau)}\mathbf{e}^{c}_{t} - \mathbf{e}^{b}_{h}\mathbf{e}^{d}_{r(\tau)}\mathbf{e}^{c}_{t} +  \mathbf{e}^{d}_{h}\mathbf{e}^{b}_{r(\tau)}\mathbf{e}^{c}_{t}) + \\
   & \ (\mathbf{e}^{a}_{h}\mathbf{e}^{d}_{r(\tau)}\mathbf{e}^{d}_{t} + \mathbf{e}^{b}_{h}\mathbf{e}^{c}_{r(\tau)}\mathbf{e}^{d}_{t} - \mathbf{e}^{c}_{h}\mathbf{e}^{b}_{r(\tau)}\mathbf{e}^{d}_{t} ) \\ 
   \label{eq16}
\end{aligned}
$}
\end{equation}

\begin{equation}
\resizebox{1.0\linewidth}{!}{$
\begin{aligned}
   & \mathbf{q}_{t} \otimes \mathbf{q'}_{r(\tau)} \cdot \mathbf{q}_h \\   
   = & \ ( - \mathbf{e}^{b}_{t}\mathbf{e}^{b}_{r(\tau)}\mathbf{e}^{a}_{h} - \mathbf{e}^{c}_{t}\mathbf{e}^{c}_{r(\tau)}\mathbf{e}^{a}_{h} - \mathbf{e}^{d}_{t}\mathbf{e}^{d}_{r(\tau)}\mathbf{e}^{a}_{h}) + \\
    & \ (\mathbf{e}^{a}_{t}\mathbf{e}^{b}_{r(\tau)}\mathbf{e}^{b}_{h} +  \mathbf{e}^{c}_{t}\mathbf{e}^{d}_{r(\tau)}\mathbf{e}^{b}_{h} - \mathbf{e}^{d}_{t}\mathbf{e}^{c}_{r(\tau)}\mathbf{e}^{b}_{h}) + \\
   & \ (\mathbf{e}^{a}_{t}\mathbf{e}^{c}_{r(\tau)}\mathbf{e}^{c}_{h} - \mathbf{e}^{b}_{t}\mathbf{e}^{d}_{r(\tau)}\mathbf{e}^{c}_{h} +  \mathbf{e}^{d}_{t}\mathbf{e}^{b}_{r(\tau)}\mathbf{e}^{c}_{h}) + \\
   & \ (\mathbf{e}^{a}_{t}\mathbf{e}^{d}_{r(\tau)}\mathbf{e}^{d}_{h} + \mathbf{e}^{b}_{t}\mathbf{e}^{c}_{r(\tau)}\mathbf{e}^{d}_{h} - \mathbf{e}^{c}_{t}\mathbf{e}^{b}_{r(\tau)}\mathbf{e}^{d}_{h}) \\ 
   \label{eq17}
\end{aligned}
$}
\end{equation}

By comparison, the above two equations are not identical because of the different signs of the terms. Therefore, when the real part of $\mathbf{q'}_{r(\tau)}$ is zero, TQuatE can model asymmetric relations.

\section{Proof of Propositions 3}
\label{sec:appendix C}
To model the inverse relation pattern, that is 
 $\forall h, t, \tau, \ r_1(h, t, \tau,) \land r_2(t, h, \tau)$ hold True accordding Definetion~\ref{def3}, which implies $\mathbf{q}_{h} \otimes \mathbf{q'}_{r_1(\tau)} \cdot \mathbf{q}_t \ = \mathbf{q}_{t} \otimes \mathbf{q'}_{r_2(\tau)} \cdot \mathbf{q}_h$. If $\mathbf{q'}_{r_2(\tau)} = \mathbf{\overline{q'}}_{r_1(\tau)}$, we have:

\begin{equation}
\resizebox{1.0\linewidth}{!}{
$
\begin{aligned}
   & \mathbf{q}_{t} \otimes \mathbf{q'}_{r_2(\tau)} \cdot \mathbf{q}_h \\   
   = & \ \mathbf{q}_{t} \otimes \mathbf{\overline{q'}}_{r_1(\tau)} \cdot \mathbf{q}_h \\
   = & \ (\mathbf{e}^{a}_{t}\mathbf{e}^{a}_{r_1(\tau)}\mathbf{e}^{a}_{h} + \mathbf{e}^{b}_{t}\mathbf{e}^{b}_{r_1(\tau)}\mathbf{e}^{a}_{h} + \mathbf{e}^{c}_{t}\mathbf{e}^{c}_{r_1(\tau)}\mathbf{e}^{a}_{h} + \mathbf{e}^{d}_{t}\mathbf{e}^{d}_{r_1(\tau)}\mathbf{e}^{a}_{h}) + \\
    & \ (\mathbf{e}^{a}_{t}\mathbf{e}^{b}_{r_1(\tau)}\mathbf{e}^{b}_{h} + \mathbf{e}^{b}_{t}\mathbf{e}^{a}_{r_1(\tau)}\mathbf{e}^{b}_{h} - \mathbf{e}^{c}_{t}\mathbf{e}^{d}_{r_1(\tau)}\mathbf{e}^{b}_{h} + \mathbf{e}^{d}_{t}\mathbf{e}^{c}_{r_1(\tau)}\mathbf{e}^{b}_{h}) + \\
   & \ (-\mathbf{e}^{a}_{t}\mathbf{e}^{c}_{r_1(\tau)}\mathbf{e}^{c}_{h} + \mathbf{e}^{b}_{t}\mathbf{e}^{d}_{r_1(\tau)}\mathbf{e}^{c}_{h} - \mathbf{e}^{c}_{t}\mathbf{e}^{a}_{r_1(\tau)}\mathbf{e}^{c}_{h} + \mathbf{e}^{d}_{t}\mathbf{e}^{b}_{r_1(\tau)}\mathbf{e}^{c}_{h}) + \\
   & \ (-\mathbf{e}^{a}_{t}\mathbf{e}^{d}_{r_1(\tau)}\mathbf{e}^{d}_{h} + \mathbf{e}^{b}_{t}\mathbf{e}^{c}_{r_1(\tau)}\mathbf{e}^{d}_{h} - \mathbf{e}^{c}_{t}\mathbf{e}^{b}_{r_1(\tau)}\mathbf{e}^{d}_{h} + \mathbf{e}^{d}_{t}\mathbf{e}^{a}_{r_1(\tau)}\mathbf{e}^{d}_{h}) \\
   = & \ (\mathbf{e}^{a}_{h}\mathbf{e}^{a}_{r_1(\tau)}\mathbf{e}^{a}_{t} - \mathbf{e}^{b}_{h}\mathbf{e}^{b}_{r_1(\tau)}\mathbf{e}^{a}_{t} - \mathbf{e}^{c}_{h}\mathbf{e}^{c}_{r_1(\tau)}\mathbf{e}^{a}_{t} - \mathbf{e}^{d}_{h}\mathbf{e}^{d}_{r_1(\tau)}\mathbf{e}^{a}_{t}) + \\
    & \ (\mathbf{e}^{a}_{h}\mathbf{e}^{b}_{r_1(\tau)}\mathbf{e}^{b}_{t} + \mathbf{e}^{b}_{h}\mathbf{e}^{a}_{r_1(\tau)}\mathbf{e}^{b}_{t} + \mathbf{e}^{c}_{h}\mathbf{e}^{d}_{r_1(\tau)}\mathbf{e}^{b}_{t} - \mathbf{e}^{d}_{h}\mathbf{e}^{c}_{r_1(\tau)}\mathbf{e}^{b}_{t}) + \\
   & \ (\mathbf{e}^{a}_{h}\mathbf{e}^{c}_{r_1(\tau)}\mathbf{e}^{c}_{t} - \mathbf{e}^{b}_{h}\mathbf{e}^{d}_{r_1(\tau)}\mathbf{e}^{c}_{t} + \mathbf{e}^{c}_{h}\mathbf{e}^{a}_{r_1(\tau)}\mathbf{e}^{c}_{t} + \mathbf{e}^{d}_{h}\mathbf{e}^{b}_{r_1(\tau)}\mathbf{e}^{c}_{t}) + \\
   & \ (\mathbf{e}^{a}_{h}\mathbf{e}^{d}_{r_1(\tau)}\mathbf{e}^{d}_{t} + \mathbf{e}^{b}_{h}\mathbf{e}^{c}_{r_1(\tau)}\mathbf{e}^{d}_{t} - \mathbf{e}^{c}_{h}\mathbf{e}^{b}_{r_1(\tau)}\mathbf{e}^{d}_{t} + \mathbf{e}^{d}_{h}\mathbf{e}^{a}_{r_1(\tau)}\mathbf{e}^{d}_{t}) \\
   = & \ \mathbf{q}_{h} \otimes \mathbf{q'}_{r_1(\tau)} \cdot \mathbf{q}_t 
   \label{eq18}
\end{aligned}
$}
\end{equation}

Therefore, when $\mathbf{q'}_{r_2(\tau)}$ is the Conjugate of $\mathbf{q'}_{r_1(\tau)}$, TQuatE can model inverse relations.

\section{Proof of Propositions 4}
\label{sec:appendix D}

To model the compositional relation pattern, that is $\forall x, y, z, \tau, r_2(x, y, \tau) \land r_3(y, z, \tau) => r_1(x, z, \tau)$ hold True accordding Definetion~\ref{def4}, which implies $(\mathbf{q}_{h} \otimes \mathbf{q'}_{r_2(\tau)}) \otimes \mathbf{q'}_{r_3(\tau)} \cdot \mathbf{q}_t = \mathbf{q}_{h} \otimes \mathbf{q'}_{r_1(\tau)} \cdot \mathbf{q}_t$. We can get

\begin{equation}
\resizebox{0.7\linewidth}{!}{$
\begin{aligned}
   & (\mathbf{q}_{h} \otimes \mathbf{q'}_{r_2(\tau)}) \otimes \mathbf{q'}_{r_3(\tau)} \cdot \mathbf{q}_t \\
   = & \ \mathbf{q}_{h} \otimes \mathbf{q'}_{r_2(\tau)} \otimes \mathbf{q'}_{r_3(\tau)} \cdot \mathbf{q}_t \\
   = & \ \mathbf{q}_{h} \otimes (\mathbf{q'}_{r_2(\tau)}) \otimes \mathbf{q'}_{r_3(\tau)}) \cdot \mathbf{q}_t \\
   = & \ \mathbf{q}_{h} \otimes \mathbf{q'}_{r_1(\tau)} \cdot \mathbf{q}_t \\
   \label{eq19}
\end{aligned}
$}
\end{equation}

Therefore, TQuatE can model compositional relations.

\section{Proof of Propositions 5}
\label{sec:appendix E}

To model the evolutionary relation pattern, that is 
$\forall h, t, \tau, r_1(h, t, \tau_1) \land r_2(t, h, \tau_2)$ hold True accordding Definetion~\ref{def5}, which implies $\mathbf{q}_{h} \otimes \mathbf{q'}_{r_1(\tau_1)} \cdot \mathbf{q}_t = \mathbf{q}_{h} \otimes \mathbf{q'}_{r_2(\tau_2)} \cdot \mathbf{q}_t$. i.e. $\mathbf{q'}_{r_1(\tau_1)} = \mathbf{q'}_{r_2(\tau_2)}$. 

We have $\mathbf{q'}_{r_1(\tau_1)} = \mathbf{q}_{\tau_1} \otimes \mathbf{q}_{r_1} \otimes \mathbf{\overline{q}}_{\tau_1} + sin(\mathbf{q}_{\tau_1'})$ and $\mathbf{q'}_{r_2(\tau_2)} = \mathbf{q}_{\tau_2} \otimes \mathbf{q}_{r_2} \otimes \mathbf{\overline{q}}_{\tau_2} + sin(\mathbf{q}_{\tau_2'})$ accordding equation~\ref{eq5} and equation~\ref{eq6}. If $sin(\mathbf{q}_{\tau_1'}) = 0$ and $sin(\mathbf{q}_{\tau_1'}) = 0$, then $\mathbf{q'}_{r_1(\tau_1)} = \mathbf{q}_{\tau_1} \otimes \mathbf{q}_{r_1} \otimes \mathbf{\overline{q}}_{\tau_1}$ and $\mathbf{q'}_{r_2(\tau_2)} = \mathbf{q}_{\tau_2} \otimes \mathbf{q}_{r_2} \otimes \mathbf{\overline{q}}_{\tau_2}$.

\begin{equation}
\resizebox{0.9\linewidth}{!}{$
\begin{aligned}
& (\mathbf{\overline{q}}_{\tau_2} \otimes \mathbf{q}_{\tau_1}) \otimes \mathbf{q}_{r_1} \otimes \overline{(\mathbf{\overline{q}}_{\tau_2} \otimes \mathbf{q}_{\tau_1})} = \mathbf{q}_{r_2} \\
\iff & \ \mathbf{\overline{q}}_{\tau_2} \otimes \mathbf{q}_{\tau_1} \otimes \mathbf{q}_{r_1} \otimes \mathbf{q}_{\tau_1} \otimes \mathbf{\overline{q}}_{\tau_2} = \mathbf{q}_{r_2} \\
\iff & \ \mathbf{\overline{q}}_{\tau_2} \otimes \mathbf{q'}_{r_1(\tau_1)} \otimes \mathbf{\overline{q}}_{\tau_2} = \mathbf{q}_{r_2} \\
\iff & \ \mathbf{q'}_{r_1(\tau_1)} = \mathbf{\overline{q}}_{\tau_2} \otimes \mathbf{q}_{r_2} \otimes \mathbf{\overline{q}}_{\tau_2} \\
\iff & \ \mathbf{q'}_{r_1(\tau_1)} = \mathbf{q'}_{r_2(\tau_2)}
   \label{eq20}
\end{aligned}
$}
\end{equation}

Therefore, TQuatE can model evolutionary relations.
\end{document}